\documentclass[11pt,a4paper]{article}

\usepackage[utf8]{inputenc}
\usepackage[T1]{fontenc}
\usepackage{lmodern}
\usepackage{microtype}
\usepackage[margin=1in]{geometry}
\usepackage{hyperref}
\usepackage{graphicx}
\usepackage{booktabs}
\usepackage{tabularx}
\newcolumntype{Y}{>{\raggedright\arraybackslash}X}
\providecommand{\seqsplit}[1]{#1}
\IfFileExists{seqsplit.sty}{\usepackage{seqsplit}}{}
\usepackage{amsmath}
\usepackage{amssymb}
\usepackage{xcolor}
\usepackage[backend=biber,style=numeric-comp,sorting=none]{biblatex}

\AtBeginBibliography{\sloppy\raggedright\setlength{\emergencystretch}{6em}}

\hypersetup{
  colorlinks=true,
  linkcolor=black,
  citecolor=blue!50!black,
  urlcolor=blue!50!black,
  pdftitle={Agentic, Context-Aware Risk Intelligence in the Internet of Value},
  pdfauthor={Basel Magableh; OmniRisk Research},
}

\title{Agentic, Context-Aware Risk Intelligence\\
       in the Internet of Value}
\author{%
  Basel Magableh\textsuperscript{1,*} \quad\quad OmniRisk Research\textsuperscript{2} \\[0.8em]
  \normalsize\textsuperscript{1}School of Computer Science, Technological University Dublin, Ireland \\
  \normalsize\textsuperscript{2}Rayachain Lab, Dublin, Ireland \\[0.4em]
  \normalsize\textsuperscript{*}Correspondence: \texttt{basel.magableh@tudublin.ie}
}
\date{May 2026}

\begin{document}
\maketitle

\begin{abstract}
The \emph{Internet of Value} (IoV) is a heterogeneous, partially-trusted
network in which the dominant marginal risk is composite (route,
sentiment, liquidity, and the policy a system is willing to commit to)
rather than a property of any single chain. We argue that a risk primitive
adequate for this regime is a \textbf{composition of five engines}: a
prediction engine over price, liquidity, volatility, and route health; a
Bittensor verification subnet that decentralises and economically scores
prediction outputs; a sentiment-fusion engine over text, on-chain flow,
and grey-literature feeds; an agentic engine under constitutional,
role-bound action constraints~\cite{bai2022constitutional}; and an
API-risk and scenario engine that converts forecasts into pre-committed
action programs in the sense of Monte-Carlo scenario
generation~\cite{glasserman2003monte}. We anchor the architecture in two
empirical artefacts: a 27-hour policy-constrained liquidity
stress-response experiment on Solana, and a 168-hour prediction-router
calibration arc reported with explicit class-imbalance honesty. The case
study supports deployability; the validator-loss decomposition is stated
formally and is falsifiable.
\end{abstract}


\smallskip
\noindent\textbf{Subject classifications.} ACM CCS: \emph{primary} ---
Security and privacy~$\to$~Distributed systems security~$\to$~Distributed
algorithms; \emph{secondary} --- Computing methodologies~$\to$~Artificial
intelligence~$\to$~Multi-agent systems; \emph{secondary} --- Applied
computing~$\to$~Electronic commerce~$\to$~Digital cash. AMS/MSC 2020:
\texttt{68M14} (distributed systems), \texttt{68T05} (learning and
adaptive systems), \texttt{91G70} (statistical methods in finance).

\section{Introduction}\label{sec:intro}

Value moves across chains. The dominant marginal risk to a participant is
now composite (route, sentiment, liquidity, and agentic policy together)
rather than a property of any single chain. The tools used to assess that
risk still think one chain at a time, and the bridge-security literature
is the closest thing to a structured response~\cite{augusto2024sok,
zhang2023sok,belchior2021survey,wu2025safeguarding}. None of that
literature, however, asks the question we ask here: \emph{what does a
context-aware risk primitive actually have to do, and how is it kept
honest?}

\paragraph{Why now.} Four failure modes that single-chain risk engines do
not capture have become individually consequential in the past two years.
\emph{Bridge fragmentation:} cross-chain bridges are now both the dominant
capital route and the dominant exploit
surface~\cite{augusto2024sok}. \emph{Liquidity fragmentation:} the same
logical asset prices differently across chains and venues, and a route that
is liquid in aggregate can be illiquid at the slice that
matters~\cite{zhang2023sok}. \emph{Narrative contagion:} sentiment shocks
propagate at a faster cadence than chain-state changes, so a risk engine
that ignores text drift is blind to a class of events whose on-chain
footprint is downstream of the
narrative~\cite{schumaker2009textual}. \emph{Agentic execution risk:} an
LLM-mediated agent that selects and executes on-chain actions adds an
attack surface that no purely on-chain mitigation
removes~\cite{bai2022constitutional}.

This paper argues that the right primitive for the IoV is a composition
of five engines (Section~\ref{sec:arch}). A \emph{prediction engine}
emits forecasts of price, liquidity, volatility, and route health. A
\emph{Bittensor verification subnet} decentralises the prediction layer
and economically scores its outputs against
realised cross-chain events~\cite{rao2021bittensor,steeves2022incentivizing,
opentensor2024yuma}. A \emph{sentiment-fusion engine} combines off-chain
text streams with on-chain flow signals using the early-fusion /
late-fusion / stacking taxonomy of the financial-NLP
literature~\cite{schumaker2009textual,xing2018natural,day2016deep}. An
\emph{agentic engine} converts forecasts and sentiment into role-bound,
constitutionally-constrained actions in the sense of Anthropic's
constitutional-AI specification~\cite{bai2022constitutional}. An
\emph{API-risk and scenario engine} turns predicted scenarios into
pre-committed action programs whose triggers and resource bounds are
declared in advance.

The architecture is grounded in two empirical artefacts. The first is a
27-hour policy-constrained liquidity stress-response experiment on a
Solana micro-cap pool (Section~\ref{sec:case-study}): 52
time-weighted-sliced buys deploying 5.2~SOL, executing under a
constitutional Trader-role contract that held across two stop-loss events
and three manual governance escalations. The second is a 168-hour
prediction-router calibration arc on the production Blue
\texttt{us-east-1} deployment (Section~\ref{sec:metrics}): empirical
accuracy improves from a 53\% live-API baseline to 99\% on a
top-cap-restricted cohort, reported with the class-imbalance and
cohort-recomposition caveats that constrain the headline. Together, the
two artefacts demonstrate the architecture is deployable and falsifiable
without yet validating the verification substrate at scale; the
validator-loss components are stated formally
(Section~\ref{sec:formal}) and remain a falsifiable target for follow-up
work~\cite{omnirisk2026soak}.

\paragraph{Contributions.} (i)~A \emph{five-engine architecture} for
agentic, context-aware IoV risk prediction (Section~\ref{sec:arch}),
composed into a single coherent dataflow with a hard architectural rule
(constitutional action constraints on the agentic engine, gated by an
upstream agent-task-management layer). (ii)~A 27-hour
\emph{policy-constrained liquidity stress-response experiment}
(Section~\ref{sec:case-study}) instrumenting the scenario engine on a
single Solana micro-cap pool as a deterministic Trader-role policy,
with 52 timestamped buys across two stop-loss events and three
out-of-band manual governance escalations. (iii)~A \emph{production
calibration arc} (Section~\ref{sec:metrics}) from the live OmniRisk
prediction router, framed honestly: the headline 99\% accuracy is
paired with the 53\% baseline, the Brier calibration error of 0.1335,
and the top-cap cohort restriction that makes the headline what it is.

\paragraph{Roadmap.} Section~\ref{sec:arch} specifies the five engines.
Section~\ref{sec:threat} states the threat model and trust assumptions.
Section~\ref{sec:formal} states the validator-loss decomposition formally.
Sections~\ref{sec:case-study}--\ref{sec:metrics} report the empirical
anchors. Section~\ref{sec:langgraph} names the production orchestration
runtime (LangGraph) and the agent-task management surface (Paperclip).
Section~\ref{sec:limits} states the falsifiable limitations.
Section~\ref{sec:conclusion} closes.

\section{The Five-Engine Architecture}\label{sec:arch}

\begin{figure}[t]
\centering
\includegraphics[width=0.92\linewidth]{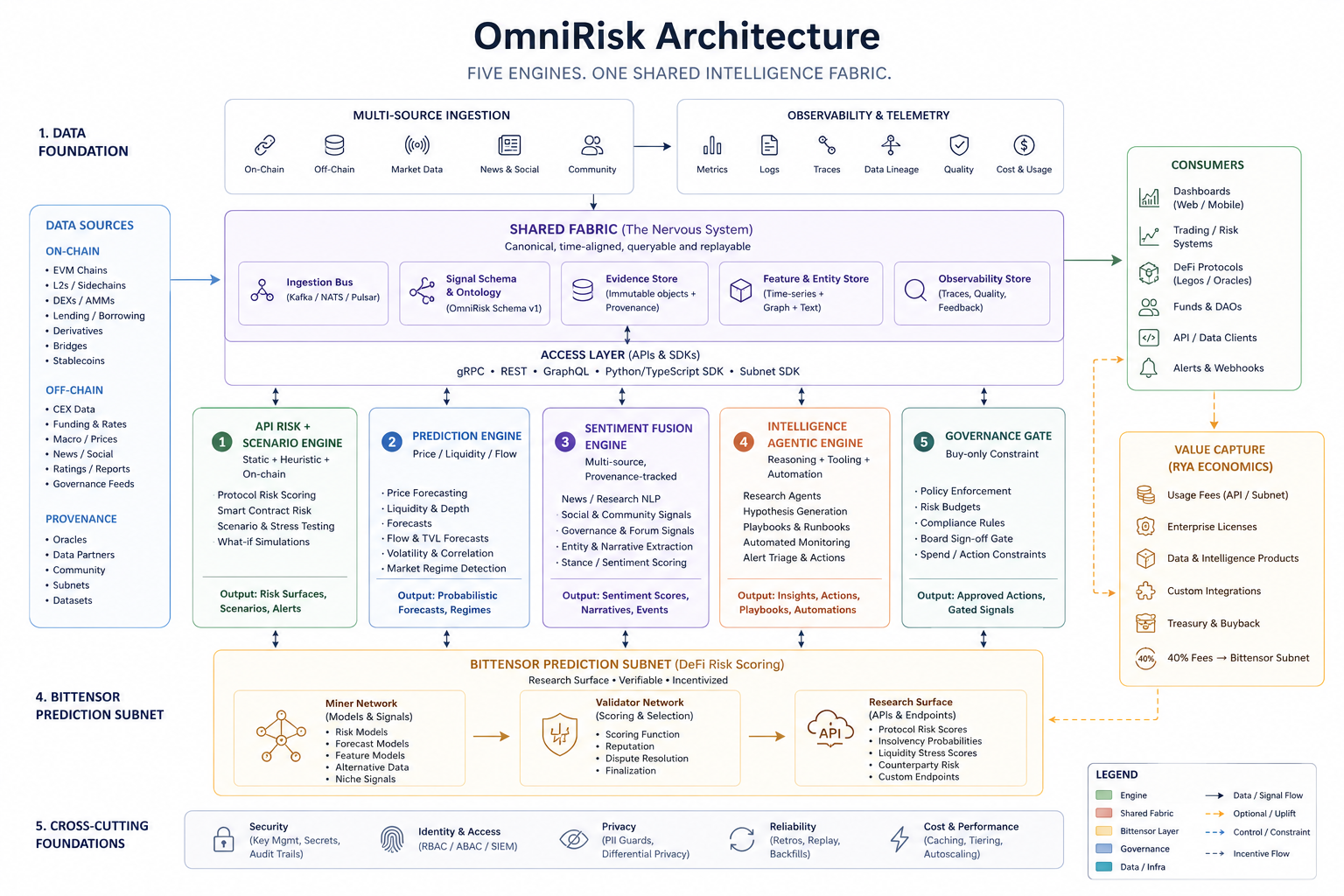}
\caption{The OmniRisk architecture: five engines composed over a single
shared intelligence fabric. Multi-source ingestion (on-chain, off-chain,
market, news, and community) and observability telemetry feed a canonical,
time-aligned, queryable, and replayable Shared Fabric (the ingestion bus,
signal schema, evidence store, feature/entity store, and observability
store), exposed through a unified access layer (gRPC, REST, GraphQL,
language SDKs, and a subnet SDK). The five engines (API-risk + scenario,
prediction, sentiment fusion, intelligence-agentic, and governance gate)
sit above the fabric and emit risk surfaces, probabilistic forecasts,
narrative-grounded sentiment, automated playbooks, and gated approvals
respectively. The Bittensor prediction subnet provides a verifiable,
incentivised research surface beneath the prediction engine, and value
capture (RYA economics) routes 40\% of fees to the subnet. Cross-cutting
foundations (security, identity-and-access, privacy, reliability, and
cost-and-performance) span every layer. Dashed amber arrows mark optional
uplifts; dashed black arrows mark control / constraint flow; dashed
incentive flow connects the value-capture surface to the subnet.}
\label{fig:omnirisk-architecture}
\end{figure}

\begin{figure}[t]
\centering
\includegraphics[width=\linewidth]{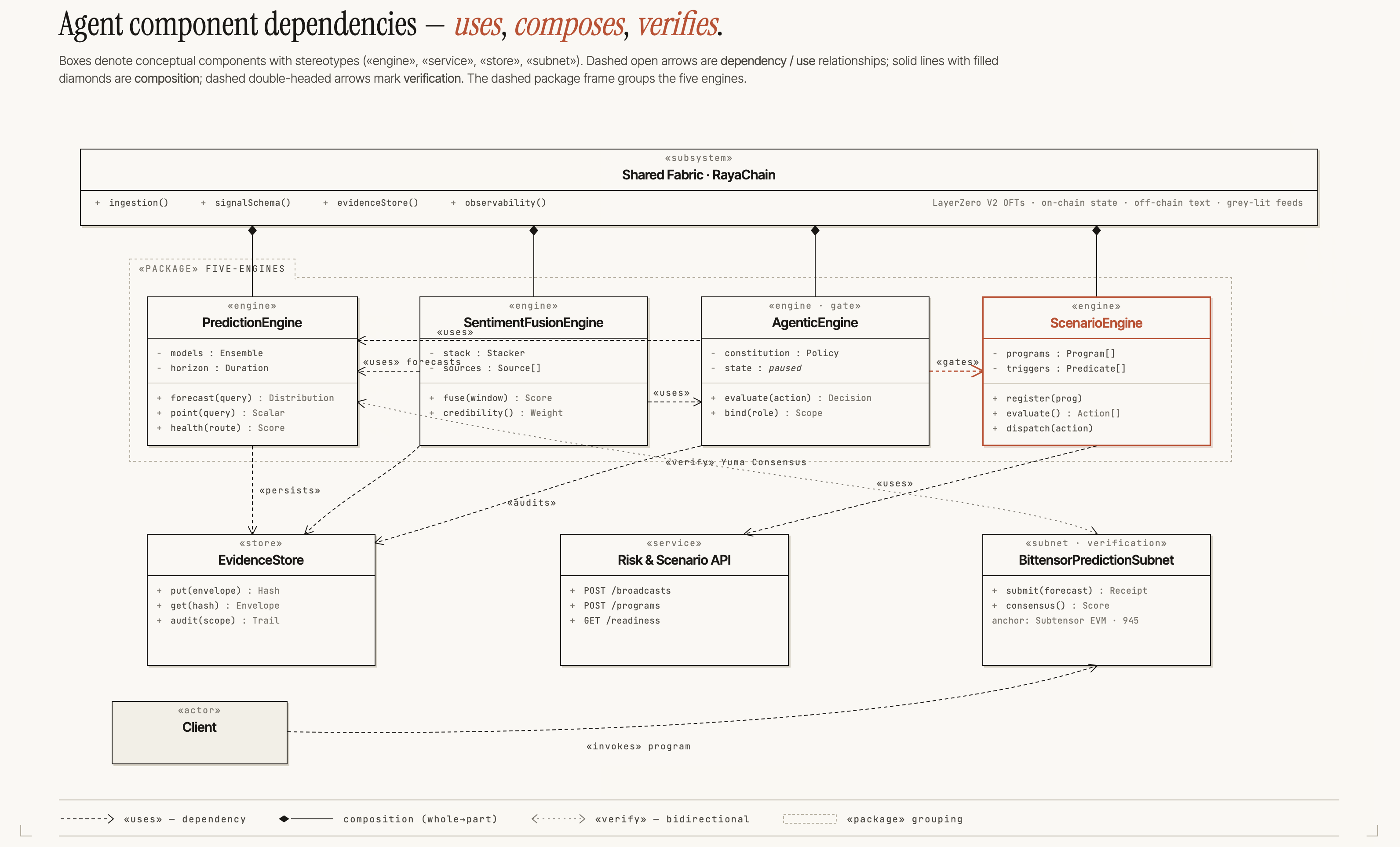}
\caption{Dependency graph for the five-engine composition (companion to
Figure~\ref{fig:omnirisk-architecture}). The shared substrate (ingestion,
signal schema, observability) feeds the API-risk + scenario, prediction,
and sentiment-fusion engines in parallel. The Bittensor prediction subnet
is a killable uplift to the prediction engine. The agentic engine is the
only true integrator and ships last; it consumes the other three as
evidence and is wrapped by a governance gate (paused on session start +
board approval).}
\label{fig:five-engine-dependency}
\end{figure}

\paragraph{Plain-English contract for the section.} A \emph{cross-chain
token} is a token that exists on multiple chains as one logical asset.
A \emph{trading pool} is a decentralised on-chain venue that quotes a
price as a function of its reserves. A \emph{calibration error} (the
Brier score) penalises both wrong predictions and overconfidence. The
\emph{Bittensor reward-allocation rule} (Yuma Consensus) takes a vector
of validator scores per miner, clips outliers against a stake-weighted
majority threshold, and pays rewards proportional to the clipped scores.
Time-weighted slicing splits a single trade into smaller pieces over a
fixed window to reduce price impact.

\subsection*{Prediction engine}

The prediction engine emits point and distributional forecasts of the
quantities that drive route- and pool-level risk: price, liquidity depth,
volatility, trade-volume distribution, and route-health scalars. It is
the input to every other engine. The OmniRisk production deployment
ships a frozen calibration manifest of 66{,}118 evaluation samples over
a 30-day prediction horizon across 21 anchor symbols, with a live-API
accuracy on the broad cohort of approximately 53\% --- characterised in
the engineering source-of-truth documentation as ``barely above
chance''~\cite{omnirisk2026systemdesign,omnirisk2026calibration}. The
architectural contribution is the verification primitive, not the
predictive performance.

\subsection*{Bittensor verification subnet}

The prediction engine is decentralised by exposing it as a Bittensor
subnet, where a heterogeneous population of miners produce forecasts and
validators score them against realised outcomes under the
reward-allocation rule~\cite{rao2021bittensor,steeves2022incentivizing,
opentensor2024yuma}. The mechanism-design lineage is the prediction-market
literature: Hanson's logarithmic market-scoring rule is the canonical
primitive that aggregates self-interested predictors into a calibrated
forecast~\cite{hanson2003combinatorial}. The closest live-2026
precedent is Taoshi's Subnet 8 (Proprietary Trading Network), which
incentivises distributed price-prediction signals~\cite{taoshi2024ptn}.
We have deployed an engineering shadow run of the OmniRisk validator
and miner pipeline against Bittensor testnet (netuid 60) since
2026-05-04; over an approximately 57-hour window the run recorded
5{,}097 successful rounds and zero authentication failures across
eleven autonomous Auth0 token refreshes~\cite{omnirisk2026soak}. The
shadow run validates the orchestration path; the validator-loss
components are deferred (see Section~\ref{sec:limits}).

\paragraph{Scoring specification (design proposal).} The verification
mechanism scores each miner $i$ at the end of a fixed scoring window
$\Delta = 24$~hours from prediction emission to realised-event resolution
(matching the prediction-router 24-hour cohort in Section~\ref{sec:metrics}).
Event resolution is binary: $y \in \{0,1\}$ indicates whether a
material chain-state change of class $E_1$--$E_4$ (defined in
Section~\ref{sec:formal}) occurs within $\Delta$. Reward lag is
prediction~$\to$~resolution~$\to$~score~$\to$~next epoch's reward
distribution; the current design carries a one-epoch lag. Confidence
calibration is binned reliability over the past $W = 1{,}000$ paired
$(c_i, y)$ samples with bin width $0.1$. These four parameters are
design proposals; this paper does not yet measure them against a live
multi-miner metagraph.

\subsection*{Sentiment-fusion engine}

A risk primitive for the IoV cannot ignore off-chain context. The
sentiment-fusion engine produces per-asset and per-route sentiment
signals, fused from a multimodal feature set: text streams (news, public
chat, posts), on-chain signals (transfers, bridge flows, governance), and
grey-literature feeds (Birdeye, DexScreener, Pump.fun for the Solana
micro-cap segment). The contemporary multimodal-sentiment surveys
distinguish three canonical strategies --- early fusion (concatenate
features), late fusion (combine per-modality model outputs), and stacking
(a learned meta-model over late-fusion outputs)~\cite{lai2023multimodal,
gandhi2023multimodal,zhang2025sentistack}. The OmniRisk production
deployment uses late fusion with stacking, and ingests
twelve named news adapters plus three social-stream
adapters~\cite{omnirisk2026systemdesign}. Verification of sentiment
outputs is performed on the same Bittensor subnet as prediction outputs.

\subsection*{Agentic engine}

The agentic engine is the LLM-mediated decision-and-action layer. Given
(forecast, sentiment, context), it selects an action from a constrained
menu and emits a structured response. The constraint surface is a
\emph{constitution} in the Anthropic sense~\cite{bai2022constitutional}:
a written principle list, rather than a free-form goal, against which
every candidate action is checked. OmniRisk's production stance is
\emph{paused-by-default}: live flips and daily-cap raises require a
board sign-off recorded in an issue thread, and the engine cannot expand
its own action surface. Each role declares its own constitution
(``buy-only, never sell'' for the Trader role; ``never use the deployer
wallet'' across all roles); the engine's action set is the intersection
of the general principles with the role's specific restrictions.

\subsection*{API-risk and scenario engine}

The scenario engine converts a forecast distribution into a
pre-committed action program of the form \emph{``if scenario~$\Sigma$
is realised in window~$W$ with probability above threshold~$p$, execute
program~$P$ within resource bound~$B$.''} This is the operational form
of a stopping rule on a Monte-Carlo sample path of the predicted joint
distribution~\cite{glasserman2003monte}. The program-registry,
trigger-evaluator, and constrained-execution decomposition follows the
NIST tabletop-exercise structure adapted to an on-chain execution
context~\cite{nist2006sp80084}. The case study of
Section~\ref{sec:case-study} runs the simplest non-trivial program in
this class.

\subsection*{Composition}

Figure~\ref{fig:five-engine-dependency} gives the dependency graph. The
shared substrate (ingestion, signal schema, evidence store, observability)
is the only hard prerequisite. The API-risk + scenario, prediction, and
sentiment-fusion engines are independent and parallelisable once the
substrate exists. The Bittensor prediction subnet is a \emph{killable}
uplift to the prediction engine: if subnet signal-to-noise is poor, the
prediction engine continues without it. The agentic engine is the only
true integrator and ships last, because it consumes the other three as
evidence and inherits the governance gate.

\subsection*{Validation status}

Table~\ref{tab:validation-status} states which engines are measured in
this paper, which are partially validated, and which are stated
architecturally but not yet measured at scale. The table preempts the
reading of the paper as a uniformly empirical artefact: only two of the
five engines have a live measurement surface here.

\begin{table}[ht]
\centering
\small
\caption{Validation status of the five engines in this paper. PARTIAL
denotes the engine is in production but not separately measured here;
SHADOW ONLY denotes a testnet-only run not yet against a real metagraph;
POLICY VALIDATED denotes the constitution and constrained-execution path
held under the case study; YES denotes a live empirical measurement
reported in this paper.}
\label{tab:validation-status}
\begin{tabularx}{\linewidth}{@{}l Y Y l@{}}
\toprule
\textbf{Engine} & \textbf{Inputs} & \textbf{Output} & \textbf{Measured?} \\
\midrule
Prediction & Market + liquidity + on-chain & Crash probability &
YES (\S\ref{sec:metrics}) \\
Sentiment fusion & Text + on-chain flow & Sentiment score & PARTIAL \\
Bittensor verification & Miner forecasts & Validator ranking &
SHADOW ONLY (testnet) \\
Agentic & Forecasts + sentiment + role contract & Policy action &
POLICY VALIDATED (\S\ref{sec:case-study}) \\
API-risk / scenario & Forecast distributions & Trigger \& bounded execution &
YES (\S\ref{sec:case-study}) \\
\bottomrule
\end{tabularx}
\end{table}

\section{Threat Model and Trust Assumptions}\label{sec:threat}

Table~\ref{tab:threat-model} states the smallest set of adversaries the
architecture is designed to tolerate, each as a row over capability,
defence, and residual risk.

\begin{table}[ht]
\centering
\footnotesize
\caption{Threat model. Each row gives the adversary's assumed
capability, the architectural defence, and the residual surface the
architecture does not cover.}
\label{tab:threat-model}
\renewcommand{\arraystretch}{1.15}
\begin{tabularx}{\linewidth}{@{}l Y Y Y@{}}
\toprule
\textbf{Adversary} & \textbf{Capability} & \textbf{Defence} &
\textbf{Residual risk} \\
\midrule
Malicious miner & Arbitrary forecast emission, adaptive to validator
queries & Three-component validator loss
(Section~\ref{sec:formal}); killable-uplift property of the subnet
& Miner that learns the query distribution may game the inconsistency
component \\
Colluding validators & Coordinated weight submission across stake
fraction $f$ & Clipped weights and stake distribution under the Yuma
rule & A $\geq 50\%$ stake coalition can still misrank; mitigated by
stake-distribution governance \\
Sentiment poisoning & Coordinated text injection across news, social,
and grey-literature feeds & Provenance-tracked ingestion;
deterministic lexicon; fusion-divergence detector vs.\ on-chain-only
baseline & Attacker that also influences on-chain flow (e.g.\ wash
trades) defeats divergence detector~\cite{cong2022crypto} \\
Oracle / route manipulation & Control over a single oracle or venue
quote, incl.\ sandwiching~\cite{daian2020flash} & Cross-source
redundancy in shared fabric; divergence flags cohort
\textsc{degraded} & Correlated manipulation across all sources not
detected \\
Replay attacker & Re-emission of past signed messages
& Window-aligned canonical timestamps; out-of-window forecasts
rejected & Same-window replay between adjacent miners is
indistinguishable from genuine convergence \\
Governance bypass & Process-level access to the runtime host
& Engine paused on session start; runtime cannot circumvent policy
boundary (no direct LLM HTTP) & Compromised host with code-execution
can still emit fabricated approval records \\
Compromised runtime node & Arbitrary code execution on the
orchestrator host & Three-tier fallback (remote $\to$ in-process
$\to$ legacy); append-only, replicated audit log & Time-to-detection
window is the unmitigated loss surface \\
\bottomrule
\end{tabularx}
\end{table}

\section{Formal Validator-Loss Specification}\label{sec:formal}\label{sec:formal:vloss}

The verification mechanism of Section~\ref{sec:arch} is stated formally
here. The specification is a design proposal: this paper does not yet
measure the loss against a live multi-miner metagraph (see
Section~\ref{sec:limits}).

\paragraph{Miner output.} A miner $i$ emits, for each query, a pair
\begin{equation}
m_i = (p_i, c_i),
\end{equation}
where $p_i \in [0,1]$ is the predicted probability of the realised event
over the scoring window~$\Delta$, and $c_i \in [0,1]$ is the miner's
self-stated calibration confidence in $p_i$.

\paragraph{Material chain-state event.} The realised event indicator
$y \in \{0,1\}$ is set to $1$ if any of four event classes occurs within
$\Delta$ of the prediction emission: $E_1$, a route-level liquidity
contraction beyond a configured threshold; $E_2$, a price drop beyond a
configured threshold over a configured sub-window; $E_3$, a bridge or
oracle anomaly registered by the cross-source redundancy check of
Section~\ref{sec:threat}; $E_4$, a governance state change recorded
on-chain. The thresholds and the union over $E_1$--$E_4$ together define
the resolver.

\paragraph{Validator loss.} The validator-side loss for miner $i$ is the
convex combination
\begin{equation}
L_i = \alpha \cdot \mathrm{Brier}(p_i, y) + \beta \cdot
\mathrm{Inconsistency}(m_i) + \gamma \cdot \mathrm{Calibration}(c_i),
\quad \alpha + \beta + \gamma = 1.
\end{equation}
The Brier component is the squared error
$\mathrm{Brier}(p, y) = (p - y)^2$ and penalises both wrong predictions
and overconfidence. The inconsistency component
$\mathrm{Inconsistency}(m_i)$ penalises drift across paired queries that
share the same chain-state context (no material change between query
times). The calibration component $\mathrm{Calibration}(c_i)$ is the
per-bin gap between stated confidence $c_i$ and realised accuracy over
the past $W$ paired $(c_i, y)$ samples (Section~\ref{sec:arch}).

\paragraph{Defaults.} We propose $\alpha = 0.6$, $\beta = 0.2$,
$\gamma = 0.2$ as the design starting point. The Brier component carries
the largest weight because it is the only component grounded directly in
realised outcomes; the inconsistency and calibration components are
intended to be hard to game without also producing accurate forecasts.
Every symbol is declared on first use: $m_i$, $p_i$, $c_i$, $y$,
$\Delta$, $\alpha$, $\beta$, $\gamma$, $W$, $E_1$--$E_4$.

\paragraph{Killable-uplift property.} The Bittensor verification subnet
is a \emph{killable uplift} to the prediction engine: if the empirical
distribution of $L_i$ across active miners has too high a variance over
a fixed validation window, the operator can disable the subnet path
without losing the production prediction surface. This bounds the
worst-case verification regression to the prediction engine's standalone
calibration error, which Section~\ref{sec:metrics} characterises.

\section{Case Study: A Policy-Constrained Liquidity Stress-Response Experiment}\label{sec:case-study}

We instrument the scenario engine on a single Solana micro-cap pool
between 2026-05-06 01:04~UTC and 2026-05-07 03:44~UTC. The experiment
has three phases. Across two stop-loss events the role contract held,
and every cap-raise was preceded by a recorded out-of-band approval
before the policy boundary widened.

\paragraph{Setup.} \textsc{RYA} is the RayaChain protocol token. On
Solana it trades on a single Raydium constant-product pool (the product
of the two reserve quantities is held constant by the pricing function,
so larger trades move the price more), launched 2026-05-05 with
37~SOL of seed liquidity. The execution wallet is a hot wallet funded
outside the treasury, Squads multisig, and deployer-key custody, and
operates under a Trader-role constitution whose hard rules are
\emph{buy-only}, \emph{never use the deployer wallet}, \emph{live mode
requires two flags plus a board sign-off recorded in the issue thread},
\emph{cap raises require a written rationale and explicit board
approval}, and \emph{no human impersonation}. The policy fires a
0.10~SOL buy (five 0.02~SOL slices over 150~s of time-weighted slicing)
when price falls below the bottom tier of a four-tier action ladder,
with a 600-second cooldown. A daily cap bounds exposure per UTC day;
a stop-loss circuit auto-pauses the engine on a 30\% drop within a
600-second window.

\paragraph{Phase I --- 2026-05-06, first stress event and policy-bound
intervention.} A stress event was registered at $T_0 = $~01:00~UTC: RYA
had drifted approximately $-60.9\%$ over the trailing 24~hours, and
pool liquidity had moved from a baseline of approximately USD~6{,}098
to USD~5{,}418. A manual governance escalation was recorded on the
issue thread, the wallet was topped up by 5.000~SOL, and the first
policy-bound buy fired at $T_0 + 4$~minutes at price
USD~$5.679\!\times\!10^{-5}$. A second escalation at $T_0 + 7$~minutes
raised the daily cap from 1.0 to 3.0~SOL with a written rationale.
Twenty-nine further buys completed at the bottom of the ladder; the
ratchet primitive raised the bottom-tier price by 50 basis points
between buys 8 and 9, and again between buys 28 and 29, when cumulative
spend crossed 2.0~SOL and 4.0~SOL respectively. By 06:13~UTC the daily
cap bound at exactly 3.000~SOL. Price recovered from
USD~$5.679\!\times\!10^{-5}$ to USD~$6.785\!\times\!10^{-5}$
($+19.5\%$); pool liquidity returned to USD~6{,}098. A first
non-policy-driven buyer printed at the reconverged price level around
10:00~UTC.

\paragraph{Phase II --- 2026-05-06/07, second stress event and ratchet
step 3.} A second stress event was registered at $T_1 = $~21:15~UTC
with magnitude $-5.84\%$ over approximately one hour, shaving liquidity
from USD~6{,}071 to USD~5{,}911. A third escalation at $T_1 + 15$
minutes raised the daily cap from 3.0 to 5.0~SOL, and the policy fired
its 31st buy at price USD~$6.591\!\times\!10^{-5}$. Twenty-one further
buys completed across the UTC date boundary; ratchet step 3 fired at
buy fifty-one (2026-05-07 00:42:18~UTC) when lifetime spend crossed
6.0~SOL. The fifty-second buy completed at 01:14:12~UTC at price
USD~$7.385\!\times\!10^{-5}$~--- the same level at which the prior
phase's first non-policy-driven buyer had printed, now bracketed by a
higher ladder.

\paragraph{Phase III --- 2026-05-07 02:38~UTC, the stop-loss event.}
A single sell pushed price USD~$0.0000769 \to 0.0000414$
($-46\%$ in approximately 3--10 minutes), draining roughly USD~1{,}666
of pool liquidity. The stop-loss circuit auto-paused the engine per
the configured 30\% drop in a 600-second window~--- exactly as designed.
\textbf{The safety circuit firing here is a feature of the architecture,
not an embarrassment.} The stop-loss exists precisely to prevent the
role contract from continuing to spend on a pool whose state has changed
faster than the policy assumed. At 03:44~UTC a manual governance
escalation was recorded on the same issue thread that authorised the
original live flip, and the engine was resumed under bounded execution
at the new USD~$0.0000414$ price floor. Daily spend at the resumption
moment was 1.46/5.0~SOL and lifetime spend was 7.205~SOL; the wallet
held 5.793~SOL post-top-up; the stop-loss circuit was re-armed at the
new liquidity baseline of approximately USD~4{,}661. The experiment
continues at the time of writing.

\paragraph{Properties demonstrated by the experiment.} Three properties
of the scenario engine running under a constitutional agentic-engine
constraint can be read directly from the run: (i)~the safety circuit
trips on its threshold, not only at design time; (ii)~the
human-in-the-loop escalation path is documented on the same surface that
authorised the original live flip, so a third-party auditor can
reconstruct the entire decision chain from the public issue thread plus
the public decision log; (iii)~the role contract holds across two
stop-loss events and three out-of-band escalations without violating any
of its hard rules. The companion operator-side
narrative~\cite{omnirisk2026defending} carries the buy-by-buy decision
log (52 rows of timestamps, prices, tier-trigger values, and slice-level
execution records).

\section{Production Calibration Arc}\label{sec:metrics}\label{sec:case-study:prediction-router}

Independent of the case study, the OmniRisk prediction engine ships a
production scheduled-evaluation harness on the Blue \texttt{us-east-1}
deployment. A Lambda fires on an EventBridge schedule, scans the
production storage layer for prediction--outcome pairs over a 168-hour
window, and persists per-cohort metrics. We report five representative
runs in Table~\ref{tab:calibration-arc}, reproduced
from~\cite{omnirisk2026predmetrics}.

\begin{table}[ht]
\centering
\small
\begin{tabular}{lrrrr}
\toprule
\textbf{Generated (UTC)} & \textbf{Window (h)} & \textbf{Samples} &
\textbf{didCrashAcc.} & \textbf{Brier} \\
\midrule
2026-04-21 02:43 (pre-fix baseline)   & 168 &  499 & 50.90\% & 0.1983 \\
2026-04-25 22:35 (post-fix)           & 168 & 1{,}393 & 77.32\% & 0.1477 \\
2026-04-26 00:01 (post-fix)           & 168 & 1{,}292 & 77.48\% & 0.1475 \\
2026-05-01 03:47 (small cohort)       & 168 &  100 & 100.00\% & 0.1314 \\
2026-05-07 01:41 (freshest)           & 168 &  915 & 99.34\% & 0.1335 \\
\bottomrule
\end{tabular}
\caption{The 168-hour calibration arc: pre-fix baseline through to the
freshest production cohort. The April~2026 live-API baseline on the
broader production cohort (not shown here) is approximately 53.15\%
(``barely above
chance''~\cite{omnirisk2026systemdesign}), against which the 168-hour
evaluation cohort is a more concentrated subset.}
\label{tab:calibration-arc}
\end{table}

\paragraph{What the headline numbers mean and do not mean.} The
99.34\% accuracy on the 2026-05-07 168-hour cohort, taken in isolation,
is suspiciously high for a binary risk classifier. Three caveats apply
simultaneously:
\begin{enumerate}
  \item \emph{Class imbalance.} The outcome resolver is bounded to the
        top 1{,}000 assets by market capitalisation~\cite{omnirisk2026systemdesign},
        a low base-rate cohort. A classifier that says ``no crash'' most
        of the time achieves a very high directional accuracy on this
        cohort almost by definition.
  \item \emph{The Brier calibration error is the metric that survives.}
        Perfect~$=$~0, marginal-50/50 prior~$=$~0.25; our
        value~$=$~0.1335 sits closer to perfect than to random but is
        well short of headline-grade.
  \item \emph{The 53\% live baseline is the right denominator.} The
        comparison that survives is against the broader production
        cohort, not the narrower 168-hour one.
\end{enumerate}
The architectural contribution of this paper is the verification
primitive --- the calibration arc is the empirical evidence that there
is a production telemetry surface against which a verification subnet
\emph{could} meaningfully score miners.

\section{Orchestration: Two Layers}\label{sec:langgraph}

A verifiable agentic system needs two distinct orchestration layers,
and this paper's contribution depends on keeping them separate. The
first is \emph{agent-task management}: the surface that scopes
\textbf{who} is allowed to act and \textbf{on what terms} --- role
contracts, the paused-by-default stance, and the requirement of an
out-of-band human approval recorded on the same audit trail that
authorises every action. The second is \emph{runtime orchestration}: a
state machine that walks each agent run through prediction, sentiment
fusion, tool invocation, and emission. The deployed instantiation uses
LangGraph for the runtime layer and a Paperclip-style task-management
service for the contract layer~\cite{omnirisk2026agentic}; the
architectural claim of this section, however, is the separation itself
and not either implementation, because conflating the two layers (one
service handling both) collapses the audit trail that makes the agentic
engine verifiable.

A single hard rule binds the runtime to the contract layer: \emph{the
runtime cannot bypass the agentic engine's policy boundary.} In our
deployment, that surfaces as a prohibition on direct LLM HTTP paths
from runtime nodes; every model invocation routes through an in-process
\texttt{AiRouter} so that model selection, fallback, daily-cost limits,
and audit logs are respected~\cite{omnirisk2026agentic}. Every emission
the runtime produces conforms to a strict JSON schema with a
\texttt{validationStatus} field in $\{\textsc{valid}, \textsc{degraded},
\textsc{fallback}\}$, written append-only to an audit log; downstream
consumers refuse any record that fails schema validation. This is a
constitutional-AI claim, not a deployment detail: a runtime that can
emit unstructured output cannot be verified against a role contract.

The runtime ships with a three-tier fallback ladder (remote orchestrator
$\to$ in-process orchestrator $\to$ legacy non-agentic path), so an
agent-infrastructure outage degrades the system to a strictly less
capable but still-correct mode rather than to a 5xx
error~\cite{omnirisk2026agentic}. The synthesis layer that composes the
five engines into a single agent decision (\emph{Strategy Alpha}, in
the engineering documentation) is the same agentic engine of
Section~\ref{sec:arch}, exposed through the runtime's terminal node;
the long-form report carries the deployment-level configuration and
test surface that this section deliberately omits.

\section{Limitations as Falsification}\label{sec:limits}

We state the load-bearing limitations as falsification statements:
what would have to be observed for the claim to fail, and the smallest
experiment that could observe it.

\begin{itemize}
  \item \emph{Single-pool case study.} A second run on a different
        Solana micro-cap pool that fails to reconverge to the
        pre-stress price under the same role contract would falsify
        the deployability claim. The smallest experiment is the same
        scenario engine pointed at any other RYA-tier asset on Solana,
        with the same buy-only constitution and a $\geq$2-cycle window.
        We make no causal claim that the policy caused reconvergence;
        the natural-experiment counterfactual is unavailable.
  \item \emph{Top-cap cohort restriction on the calibration arc.} The
        99.34\% directional accuracy is on a top-1{,}000-by-mcap
        cohort. The smallest experiment that would falsify the
        broad-cohort claim is to lift the cohort restriction and re-run
        the 168-hour evaluation; the hypothesis would be falsified if
        the broad-cohort calibration error climbs above 0.20.
  \item \emph{Validator-loss components are not yet measured at
        scale.} The Bittensor shadow soak validates the
        orchestration path, but the three components of the validator
        loss (Brier accuracy + Inconsistency over paired queries with
        no material chain-state change + Miscalibration via reliability
        bins) are deferred~\cite{omnirisk2026soak}. The smallest
        experiment that would break the verification claim is to
        introduce a second miner with a non-trivial protocol, accumulate
        $(c, y)$ pairs across realised events, and report the
        three-component decomposition in a follow-up. Until that runs,
        the verification claim is architectural, not measured.
\end{itemize}

A fourth honest disclosure: the agentic engine in the case study is a
deterministic Trader-role policy, not the LLM-mediated agentic engine of
Section~\ref{sec:arch}. The case study validates the
program-registry / trigger-evaluator / constrained-execution
decomposition; the LLM-mediated agentic engine is a follow-on build.

\section{Conclusion}\label{sec:conclusion}

We have proposed a five-engine composition for agentic, context-aware
risk prediction in the Internet of Value, anchored to two empirical
artefacts: a 27-hour policy-constrained liquidity stress-response
experiment whose role contract held across two stop-loss events and
three out-of-band escalations, and a 168-hour production calibration arc
reported with the class-imbalance and cohort-recomposition caveats that
prevent the headline from being misread as a standalone accuracy.
Falsifiable hypotheses H1
(cross-chain liquidity-fragmentation features improve early-detection
lead time over single-chain features) and H2
(confidence-calibrated miners receive monotonically higher Yuma weight
than uncalibrated ones) are stated formally there, and the smallest
experiments that would falsify each are listed in
Section~\ref{sec:limits} above.

\section*{Author contributions}

We use the CRediT taxonomy. \textbf{Basel Magableh:} conceptualisation,
methodology, funding acquisition, project administration, supervision,
writing --- original draft, writing --- review and editing.
\textbf{OmniRisk Research:} software, investigation, data curation,
resources, visualisation, validation, writing --- review and editing.
ORCID: Basel Magableh, \texttt{0000-0003-2337-637X}
(\url{https://orcid.org/0000-0003-2337-637X}).

\section*{Funding}

This work received no external research funding. The on-chain
stress-response experiment described in Section~\ref{sec:case-study}
was funded by the corresponding author from personal non-treasury,
non-Squads, non-deployer reserves; specific amounts and dates are
documented in the public decision log cited
as~\cite{omnirisk2026defending}. The OmniRisk production infrastructure
used for the Section~\ref{sec:metrics} calibration arc is operated by
Rayachain Lab.

\section*{Data availability}

All numerical results in Section~\ref{sec:case-study} are reproducible
from the public decision log~\cite{omnirisk2026defending} and from an
immutable CSV snapshot of the prediction-router metrics committed
alongside this manuscript at
\seqsplit{papers/2026/omnirisk-rayachain-bittensor/sources/prediction-router-metrics-2026-05-07.csv}
(commit \texttt{b419338}). The Bittensor shadow-validator soak
telemetry of Section~\ref{sec:arch} is similarly committed at
\seqsplit{papers/2026/omnirisk-rayachain-bittensor/sources/bittensor-shadow-soak-2026-05-07.md}
(commit \texttt{e726153}). The live OmniRisk prediction-router endpoint
is operator-only by design, consistent with the access-control posture
argued in the long-form companion report; reviewer access can be
requested under NDA.

\section*{Conflict of interest}

The authors declare no competing financial interests other than those
disclosed in the funding statement above (the corresponding author is
the operator of the wallet that funded the
Section~\ref{sec:case-study} experiment).

\section*{Ethics}

No human-subjects research; no PII processing; no third-party data
acquisition under restricted licence. All on-chain data are publicly
observable. The experimental policy operated on the corresponding
author's non-treasury wallet only and did not act on any other party's
funds. The engine's governance constraints (paused-by-default,
board-sign-off-required for live flips and cap raises, hard-armed
stop-loss) are documented in the public role contract cited
as~\cite{omnirisk2026defending}.

\printbibliography

\end{document}